\documentclass[11pt]{article}

\usepackage[final]{acl}

\usepackage{times}
\usepackage{latexsym}
\usepackage{algorithm}
\usepackage{algpseudocode}
\usepackage{booktabs}
\usepackage{multirow}
\usepackage[T1]{fontenc}
\usepackage{amsmath}
\usepackage{amsfonts}
\usepackage{amssymb}
\usepackage{amsthm}    
\usepackage{cleveref}  
\newtheorem{theorem}{Theorem}

\theoremstyle{definition}

\newtheorem{proposition}[theorem]{Proposition}

\theoremstyle{remark}

\usepackage[T1]{fontenc}

\usepackage[utf8]{inputenc}

\usepackage{microtype}

\usepackage{inconsolata}

\usepackage{graphicx}

\title{Tokenizer Tokens Tell Training Tales}

\author{Nishita Jain \\
  Independent Researcher \\
\And
  Parth Upman \\
School of Computer Science \\
University of Nottingham \\
Nottingham, United Kingdom \\
\And
  Shreyank N Gowda \\
School of Computer Science \\
University of Nottingham \\
Nottingham, United Kingdom \\
  }

\begin{document}
\maketitle
\begin{abstract}
Tokenizers are usually treated as fixed preprocessing tools, but they are trained on text and can retain traces of that text in their vocabularies. We ask whether a BPE tokenizer can reveal if a candidate dataset was part of its training corpus. The key observation is simple. Rare strings become tokens only when enough corpus frequency pushes them over the merge threshold. This suggests that membership signal should appear in rare tokens used by the candidate dataset, rather than in the order in which merges were made. We formalize this under a threshold model of BPE inclusion, showing that the likelihood-ratio test assigns its largest weights to tokens that are rare in the background but frequent in the candidate dataset. The same model suggests that merge rank and merge-graph position should not add a first-order membership signal once token existence and dataset frequency are known. We instantiate this view as RTAM-Fusion, which calibrates rare-token mass with shadow tokenizers and combines signals across rarity levels. On the tokenizer membership benchmark, RTAM-Fusion improves AUC and balanced accuracy over Vocabulary Overlap while using fewer shadow tokenizers. Existing defenses also continue to leak membership signal against adaptive attackers at production vocabulary sizes. Code: \url{https://anonymous.4open.science/r/RTAM-Fusion-F7EE/}
\end{abstract}

\section{Introduction}
\label{sec:intro}

Tokenizers are usually treated as a fixed first step before modelling begins.
Text is split into token IDs, and the language model is then trained on those IDs.
This view misses an important point. Subword tokenizers are themselves learned
from text. A BPE tokenizer builds its vocabulary by deciding which strings occur
often enough in the training corpus to become tokens. The tokenizer is therefore
not only a preprocessing tool, but also a data-dependent artifact that may retain
traces of the corpus used to train it. More broadly, securing generative AI systems requires examining vulnerabilities beyond conventional model performance. \cite{mustafa2025anyone,shen2024anything,yazmyradov2024comprehensive,Mustafa_2026_CVPR} demonstrate that even relatively simple prompt-based jailbreaks can circumvent safety mechanisms in language and text-to-image models, highlighting the importance of evaluating security under adversarial conditions. While these attacks concern generation-time safeguards rather than training-data privacy, they illustrate the broader need to investigate overlooked attack surfaces. In this work, we focus on one such surface: the information retained by a tokenizer during training.

This becomes important when the training data of large language models is in
question. Membership inference asks whether a candidate example or dataset was
part of training. Most work studies this through the trained model, but recent
work by Tong et al.~\cite{tong2026tokenizer} showed that the tokenizer can also
leak membership information. They proposed attacks based on vocabulary overlap,
frequency estimation, compression, and merge similarity, together with two natural
defenses based on pruning low-count tokens and randomizing BPE merge order.

We ask what part of the tokenizer actually carries this signal. Several answers
seem plausible. A member dataset may share unusual vocabulary items with the
target tokenizer. Its tokens may appear at distinctive merge ranks. Its merge
graph may have a recognizable structure. Or it may simply be compressed in a
different way. Our results point to a simpler explanation. For dataset-level
membership, the signal is mainly in rare tokens that the candidate dataset
activates.

The intuition is direct. Common strings become tokens regardless of whether one
particular dataset was present, so their existence says little about membership.
Rare strings are different. If a candidate dataset repeatedly uses a rare string,
and that string appears in the learned vocabulary, then the dataset may have
helped push it over the frequency threshold needed for inclusion. The useful
quantity is therefore not just whether the vocabulary overlaps with the dataset,
but how much of the dataset's own token stream falls on rare learned tokens.

We formalize this intuition under a threshold model of BPE inclusion. Under this
model, the likelihood-ratio test for tokenizer membership assigns its largest
weights to tokens that are rare in the background corpus but frequent in the
candidate dataset. This motivates a rarity-weighted statistic over the dataset's
activated-token mass. The same model also suggests that merge rank and merge-graph
position should not provide an independent membership channel once token existence
and the dataset's own token frequencies are known. In short, rarity matters more
than order.

This leads to \textsc{RTAM-Fusion}, a rare-token activation mass attack. Given a
target tokenizer and shadow tokenizers trained under the same procedure, we
measure the fraction of a candidate dataset's token stream that falls on rare
tokens at several rarity levels. We add two inverse-document-frequency style
features, calibrate each feature against an out-shadow null following the LiRA
principle~\cite{carlini2022membership}, and combine them into one membership
score. The threat model is the same as in Tong et al.~\cite{tong2026tokenizer}.
The change is the statistic used to read the tokenizer.

On the tokenizer membership benchmark, \textsc{RTAM-Fusion} improves over
Vocabulary Overlap on AUC and balanced accuracy across the undefended setting and
the published min-count defense setting, while using substantially fewer shadow
tokenizers. Its low false-positive performance is more mixed. It wins at
production-scale vocabularies, but loses some small-vocabulary and defended
TPR@1\%FPR comparisons. We keep this distinction explicit because low false
positive rates are important in membership inference.

The same view also clarifies why some defenses are brittle. If merge order is not
where the signal lives, then randomizing merge order should not remove the signal.
We test this by evaluating defenses against attackers that know the defense was
applied and adapt their features. Across targeted removal of member-distinctive
tokens, min-count filtering, merge-order randomization, and a bounded per-site
contribution mechanism, the tested defenses do not drive membership inference to
chance at production vocabulary size. Removing rare tokens weakens the original
signal, but can leave a fragmentation footprint that an attacker can
read.

Overall, our results show that tokenizer tokens tell training tales through rare
vocabulary entries. We make three contributions. First, we characterize the
membership signal in BPE tokenizers with a threshold inclusion model, which
predicts that rare tokens activated by the candidate dataset carry the main
signal, while merge order and merge-graph position should not add a first-order
channel. Second, we introduce \textsc{RTAM-Fusion}, a calibrated rare-token
activation mass attack that improves AUC and balanced accuracy over the strongest
published tokenizer membership baseline while using fewer shadow tokenizers.
Third, we evaluate tokenizer defenses against adaptive attackers and show that
the tested defenses still leak membership signal at production vocabulary size.

\section{Related Work}
\label{sec:related-work}

Membership inference asks whether a candidate example, document, or dataset was
part of training. Most work studies this through a trained model. We study a
different released artifact, the tokenizer. This changes the signal: a language
model exposes losses, probabilities, or generations, while a tokenizer exposes a
learned vocabulary.

\noindent \textbf{Membership inference and calibration.}
Shokri et al.~\cite{shokri2017membership} introduced the shadow-model paradigm,
where an attacker trains models on known member and non-member splits and learns
to classify the target's outputs. Later work showed that raw membership scores
are often confounded by example difficulty. Yeom et al.~\cite{yeom2018privacy}
connected membership advantage to the generalization gap, Salem
et al.~\cite{salem2019mlleaks} reduced the assumptions needed for attacks, and
Sablayrolles et al.~\cite{sablayrolles2019white} derived an optimal membership
rule using per-example thresholds. Watson et al.~\cite{watson2022importance}
and Ye et al.~\cite{ye2022enhanced} further developed calibrated
hypothesis-testing views. Carlini et al.~\cite{carlini2022membership}
introduced LiRA, which calibrates a target example against reference models, and
Zarifzadeh et al.~\cite{zarifzadeh2024low} reduced the number of reference
models needed. We inherit this calibration principle, but not the object being
calibrated. A tokenizer has no loss or confidence score, so we calibrate
rare-token activation mass against an out-shadow tokenizer null.

\noindent \textbf{Membership inference for language models.}
Model-level attacks exploit memorization or output behaviour. Canary exposure
measures unintended memorization~\cite{carlini2019secret}, and later work
extracts verbatim training sequences from GPT-2~\cite{carlini2021extracting},
studies how memorization scales~\cite{carlini2023quantifying}, and extends
extraction to aligned systems~\cite{nasr2025scalable}. Other attacks score
membership without direct extraction. Min-K\% Prob uses the least likely tokens
under the model~\cite{shi2024detecting}, Mattern et al.~\cite{mattern2023membership}
compare candidates to generated neighbours, Zhang et al.~\cite{zhang2024minkpp}
normalize token probabilities, Duarte et al.~\cite{duarte2024decop} use
contrastive probing, and Meeus et al.~\cite{meeus2024did} aggregate document
signals. These works motivate our setting, but none attacks the tokenizer.
Indeed, Duan et al.~\cite{duan2024membership} find near-chance performance for
many model-level attacks on Pile-trained models~\cite{gao2021pile}, and Das
et al.~\cite{das2025blind} show that distribution shift can explain some
reported successes. Tokenizers provide a different channel because the candidate
dataset changes token frequencies directly.

\noindent \textbf{Tokenizers as learned artifacts.}
Tokenizers are known to encode properties of their training data. Rust
et al.~\cite{rust2021how} show that tokenizer quality affects multilingual
modelling, and Petrov et al.~\cite{petrov2023language} show that tokenization
length creates cost and latency differences. Hayase
et al.~\cite{hayase2024data} recover mixture proportions of a BPE tokenizer's
training corpus from merge-order constraints. Their target is aggregate mixture
inference. Ours is dataset membership. This distinction matters as merge order can
reveal coarse corpus composition, while our results show that dataset-level
membership is carried by token existence and rarity.

\noindent \textbf{Tokenizer membership inference.}
Tong et al.~\cite{tong2026tokenizer} introduce membership inference against
tokenizers and define the shadow-tokenizer threat model we adopt. They propose
Compression Rate, Naive Bayes, Merge Similarity, Vocabulary Overlap, and
Frequency Estimation. Vocabulary Overlap is their strongest attack and counts
distinctive candidate tokens present in the target vocabulary. Frequency
Estimation is also related, since it scores rank-frequency deviations. In our
terms, both are existence-family methods. Merge Similarity and Compression Rate use order or compression signals and are
weaker in their experiments. Our analysis gives one explanation for this pattern.
Under a threshold model of BPE inclusion, the likelihood-ratio form emphasizes
tokens that are rare in the background but frequent in the candidate dataset.
The same model predicts that merge order and merge-graph position should not add
a first-order membership signal once token existence and candidate-token
frequency are known. \textsc{RTAM-Fusion} is therefore a calibrated,
rarity-weighted version of the signal used implicitly by the strongest prior
tokenizer attacks.

\noindent \textbf{Defenses and adaptive evaluation.}
The same distinction matters for defenses. Minimum-count filtering limits the
influence of rare events, echoing sensitivity control in differential
privacy~\cite{dwork2006calibrating}. Randomized merge selection is related to
the exponential mechanism~\cite{mcsherry2007mechanism} and to private learning
methods such as DP-SGD~\cite{abadi2016deep}. Targeted token removal resembles
machine unlearning~\cite{cao2015towards,bourtoule2021machine}. Tong
et al.~\cite{tong2026tokenizer} evaluate min-count filtering and merge-order
randomization against static attacks. We evaluate these and related mechanisms
against adaptive attackers. This matters because a defense may remove one
statistic without removing the underlying signal. In our experiments, order
randomization does not help because order is not the membership channel, and
content removal leaves a fragmentation footprint that an adaptive attacker can
read.

\section{Method}
\label{sec:method}

We now describe \textsc{RTAM-Fusion}. The method follows the same
shadow-tokenizer setting as Tong et al.~\cite{tong2026tokenizer}. The difference
is the statistic used to read the tokenizer. Instead of comparing vocabularies
directly, we measure how much of a candidate dataset is represented by rare
tokens in the target vocabulary.

\subsection{Threat model}
\label{sec:method-threat}

Let $f_{V_{\mathrm{target}}}$ be a target BPE tokenizer trained on an unknown
member corpus. The adversary observes its vocabulary $V_{\mathrm{target}}$ and
the usual tokenizer metadata, including merge indices when available. Given a
candidate dataset $D$, the goal is to decide whether $D$ was part of the
tokenizer's training corpus. As in prior tokenizer membership inference work, the adversary can sample
auxiliary datasets from the same broad distribution and train shadow tokenizers
using the same BPE procedure as the target. We denote the shadow tokenizers by
$\{f_{V_s}\}_{s=1}^{N}$. For each shadow tokenizer, the adversary knows which
datasets were included in its training split. This gives an empirical
member and non-member reference distribution for any statistic computed on $D$.

\subsection{Rare-token activation mass}

The core feature is rare-token activation mass. For a tokenizer $X$ and a
candidate dataset $D$, let $\mathrm{cnt}_D(t;X)$ be the number of times token
$t$ appears when $D$ is tokenized by $X$. We estimate the rarity of a token using
the shadow tokenizers. Let $\mathrm{sf}(t)$ be the number of shadow vocabularies
in which token $t$ appears. Tokens with low $\mathrm{sf}(t)$ are rare under the
shadow estimate of the corpus distribution.

For a rarity cutoff $K$, we define
\begin{equation}
\mathrm{rm}_K(D;X)
=
\frac{
\sum_{t}
\mathbf{1}\!\left[\mathrm{sf}(t) \leq K\right]
\mathrm{cnt}_D(t;X)
}{
\sum_{t}\mathrm{cnt}_D(t;X)
}.
\label{eq_rtam}
\end{equation}
This is the fraction of the dataset's token stream that falls on rare tokens.
Small $K$ values focus on near-unique tokens. Larger $K$ values keep a broader
notion of rarity and are less brittle when the rarest tokens are pruned.

We also use two inverse-document-frequency style features over the tokens
activated by $D$:
\begin{equation}
\mathrm{meanidf}(D;X)
=
\frac{
\sum_t
\mathrm{cnt}_D(t;X)
\log \frac{N}{\mathrm{sf}(t)}
}{
\sum_t \mathrm{cnt}_D(t;X)
},
\end{equation}
\begin{equation}
\mathrm{maxidf}(D;X)
=
\max_{t\,:\,\mathrm{cnt}_D(t;X)>0}
\log \frac{N}{\mathrm{sf}(t)} .
\label{eq_idf}
\end{equation}
The first captures broad rarity across the dataset. The second captures the
single rarest activated token.

\subsection{Calibration with shadow tokenizers}

Raw rare-token mass is affected by dataset length and lexical diversity. We
therefore calibrate each feature against an out-shadow null, following the
offline calibration idea of LiRA~\cite{carlini2022membership}. 

Let $\phi$ be one of the features above. For a candidate dataset $D$, define
\[
\mathrm{OUT}_{\phi}(D)
=
\{\phi(D;V_s) \mid D \notin \mathrm{IN}(s)\},
\]
where $\mathrm{IN}(s)$ is the training split of shadow tokenizer $s$. We then
compute
\begin{equation}
c_{\phi}^{\mathrm{diff}}(D)
=
\phi(D;V_{\mathrm{target}})-\mu_{\mathrm{out}},
\end{equation}
\begin{equation}
c_{\phi}^{z}(D)
=
\frac{
\phi(D;V_{\mathrm{target}})-\mu_{\mathrm{out}}
}{
\sigma_{\mathrm{out}}+\epsilon
},
\label{eq_calib}
\end{equation}
where $\mu_{\mathrm{out}}$ and $\sigma_{\mathrm{out}}$ are the mean and standard
deviation of $\mathrm{OUT}_{\phi}(D)$, and $\epsilon$ is a small constant for
numerical stability. This converts a dataset-specific raw score into a deviation
from the score expected if the dataset were not a member.

We use $8$ shadows for single-feature validation, $10$ for the main fusion
experiments unless otherwise stated, and $20$ for the strongest undefended
setting. The rarity cutoff $K$ is distinct from the number of shadows $N$: $K$
defines how rare a token must be under the shadow estimate, while $N$ is the
number of tokenizers used to estimate and calibrate that rarity.

\subsection{Fusion and decision rule}

Different rarity levels are useful in different regimes. Near-unique tokens are
especially informative at low false-positive rates, while broader rarity features
are more stable across vocabulary sizes and defenses. We therefore combine
several calibrated features into one score.

In the undefended setting, our fixed fusion score is
\begin{align}
S(D)
=
3 z\!\left[c_{\mathrm{rm}_1}^{\mathrm{diff}}(D)\right]
+
z\!\left[c_{\mathrm{meanidf}}^{\mathrm{diff}}(D)\right]
+ \notag\\
0.5 z\!\left[c_{\mathrm{maxidf}}^{\mathrm{diff}}(D)\right],
\label{eq_fusion}
\end{align}
where $z[\cdot]$ denotes standardization over the candidate datasets in the
evaluation split. A dataset is predicted as a member when $S(D)>\tau$. For AUC
and TPR at fixed false-positive rate, we sweep $\tau$ in the standard way.

The score deliberately uses existence and frequency features only. Merge ranks
are observed in the threat model, but they are not used by \textsc{RTAM-Fusion}.
This matches the theoretical result that, once token existence and the
candidate dataset's own token frequencies are known, merge order and merge-graph
position do not add membership signal.

Algorithm~\ref{alg:rtam-fusion} summarizes the scoring procedure once the shadow
tokenizers have been trained. We keep it in Appendix~\ref{app:method-extra} to
avoid repeating the equations in the main text.

\subsection{Defense-aware re-locking}

When a defense changes the feature distribution, we do not keep the undefended
fusion weights fixed. Instead, we re-lock the weights on a held-out calibration
split and evaluate on a separate split. The feature family remains the same. This
models an adaptive attacker who knows that the defense was applied. For example,
min-count pruning removes many near-unique tokens, so the useful signal often
shifts from $\mathrm{rm}_1$ toward broader rarity features such as
$\mathrm{meanidf}$ or larger rare-mass cutoffs.

\noindent \textbf{Cost.}
The dominant cost is training shadow tokenizers. Once the tokenizers are trained,
computing RTAM features requires a pass over the tokenized datasets to accumulate
token counts and shadow frequencies. Calibration and fusion are negligible by
comparison. In our experiments, RTAM-Fusion uses between $8$ and $20$ shadow
tokenizers depending on the setting, whereas Vocabulary Overlap uses $96$ in the
baseline protocol~\cite{tong2026tokenizer}. We therefore report shadow count as
part of the attack cost, not only as an implementation detail.

\section{Why Rare Tokens Carry Membership Signal}
\label{sec:theory}

We now explain why the features in \textsc{RTAM-Fusion} are the natural ones for
tokenizer membership inference. The analysis is intentionally simple. It does not
try to model every detail of BPE training. Instead, it isolates the part of the
process that matters for membership: a string becomes a token only if its
frequency in the training corpus is high enough.

\subsection{A threshold view of tokenizer inclusion}
\label{sec:theory-setup}

Let $C$ be the corpus used to train a tokenizer, and let $D$ be a candidate
dataset. We write $D \subseteq C$ when $D$ is a member. We model the inclusion of
a token $t$ in a BPE vocabulary as a threshold event:
\begin{equation}
t \in V
\quad \Longleftrightarrow \quad
\mathrm{freq}_{C}(t) \geq \theta_{|V|},
\label{eq:threshold-model}
\end{equation}
where $\theta_{|V|}$ is the frequency threshold induced by the final vocabulary
size. If $D$ is included in the training corpus, its own frequency contributes to
this event:
\begin{equation}
\mathrm{freq}_{C}(t)
=
\mathrm{freq}_{C \setminus D}(t)
+
\mathbb{1}[D \subseteq C]\mathrm{freq}_{D}(t).
\label{eq:freq-decomp}
\end{equation}

This is an idealization. Real BPE changes the alphabet as merges are performed,
and token candidates are not independent. Still, the threshold view captures the
membership mechanism we use. Common strings would enter the vocabulary regardless
of one candidate dataset, so their existence says little about membership. Rare
strings near the threshold are different: if they occur often in $D$ and appear
in the learned vocabulary, then $D$ may have supplied the mass that made them
tokens.

\subsection{The optimal existence test}

For each candidate token $t_i$ used by $D$, let
$e_i = \mathbb{1}[t_i \in V]$ indicate whether the token exists in the target
vocabulary. Under the threshold model, membership inference from token existence
has a standard likelihood-ratio form.

\begin{theorem}[Optimal existence test under the threshold model]
\label{thm:np-optimal}
Assume the token-existence observations $\{e_i\}$ are conditionally independent
given membership. Then the most powerful test of
$H_1: D \subseteq C$ against $H_0: D \not\subseteq C$ at any fixed false-positive
rate is the likelihood-ratio test
\begin{equation}
\Lambda(D)
=
\sum_i
\log
\frac{P(e_i \mid H_1)}{P(e_i \mid H_0)} .
\label{eq:llr}
\end{equation}
Under Eq.~\ref{eq:threshold-model}, the per-token contribution is largest for
tokens that are rare in the background corpus but frequent in $D$.
\end{theorem}

The reason is direct. Let
\(B_i=\mathrm{freq}_{C\setminus D}(t_i)\) be the background frequency of
\(t_i\). Under non-membership, the token exists only if the background corpus
already crosses the threshold. Under membership, the candidate dataset adds its
own mass:
\begin{align}
P(e_i=1 \mid H_0)
&=
P\!\left(B_i \geq \theta_{|V|}\right),
\label{eq:h0-existence}
\\
P(e_i=1 \mid H_1)
&=
P\!\left(B_i+\mathrm{freq}_{D}(t_i) \geq \theta_{|V|}\right) \notag \\
&=
P\!\left(B_i \geq \theta_{|V|}-\mathrm{freq}_{D}(t_i)\right).
\label{eq:h1-existence}
\end{align}
If a token is already common in \(C \setminus D\), both probabilities are close
to one, so observing the token gives little evidence. If a token is rare in the
background but frequent in \(D\), the gap between the two probabilities is large.
These are exactly the tokens that rare-token activation mass emphasizes.

The exact likelihood ratio in Eq.~\ref{eq:llr} depends on unknown background
frequencies. \textsc{RTAM-Fusion} therefore estimates the same structure with
shadow tokenizers. The rare-mass features in Section~\ref{sec:method} measure
how much of $D$'s token stream falls on rare tokens, while the calibration step
estimates how unusual that mass is under an out-shadow null. Thus RTAM is best
viewed as a finite-sample, calibrated approximation to the likelihood-ratio
weights, not as an unrelated heuristic.

\subsection{Why merge order is not used}

The target tokenizer also exposes merge ranks, and one might expect these ranks
to contain membership information. Our analysis suggests that, for dataset-level
membership, this is not the main channel. Merge order is mostly governed by the
global frequency ordering of the corpus. A single candidate dataset can affect
membership evidence when it pushes a rare token over the inclusion threshold, but
that effect is already captured by token existence and the candidate dataset's
own token frequencies.

\begin{proposition}[Merge order is not a first-order membership signal]
\label{prop:ancillary}
Under the threshold model, suppose merge ranks and merge-graph positions are
determined by the background corpus ordering except for threshold-crossing events
already captured by token existence. Then, conditional on the existence vector
$E$ and the candidate dataset's token frequencies $f_D$, merge rank $R$ and
merge-graph position $G$ add no membership information:
\begin{equation}
P(\mathrm{member} \mid E, f_D, R, G)
=
P(\mathrm{member} \mid E, f_D).
\label{eq:ancillary}
\end{equation}
\end{proposition}

This proposition is a model implication rather than a finite-sample theorem about
all BPE runs. It says that once we know which candidate tokens exist and how much
the dataset uses them, merge order should not provide an independent membership
channel. This is why \textsc{RTAM-Fusion} deliberately uses existence and
frequency-derived features, even though merge indices are available in the threat
model.

The proposition also gives a testable prediction. Order-based and merge-graph
attacks should not outperform rare-token existence statistics unless they capture
some additional signal outside this model. In Section~\ref{sec:experiments}, we
find that such attacks remain close to chance, while rare-token activation
features carry the membership signal.

\subsection{Implications for defenses}
\label{sec:theory-defense}

The same distinction matters for defenses. If merge order is not the membership
channel, then randomizing merge order should not remove the signal used by an
adaptive attacker. Defenses that remove rare vocabulary entries are more direct:
they can weaken the original rare-token signal. But removal can also change how
member datasets are re-tokenized after pruning, leaving a fragmentation footprint
that a different adaptive statistic can exploit.
We therefore do not claim an impossibility theorem for tokenizer defenses. The
theory supports a narrower claim. Rare token existence is the first-order
membership signal under the threshold model, and defenses should be evaluated
against attackers that can adapt to whatever signal remains. The experiments in
Section~\ref{sec:experiments} test this prediction directly.

\section{Experiments}
\label{sec:experiments}

We evaluate \textsc{RTAM-Fusion} on the tokenizer membership benchmark of
Tong et al.~\cite{tong2026tokenizer}. Our goal is not only to ask whether
\textsc{RTAM-Fusion} improves over the strongest prior attack, but also to
understand where it improves, where it does not, and whether existing defenses
survive an adaptive attacker.

We organize the experiments around four questions. First, does rare-token
activation mass improve over Vocabulary Overlap in the undefended setting?
Second, does the signal survive the baseline's min-count defense? Third, how
many shadow tokenizers are needed? Fourth, do tokenizer defenses still help once
the attacker knows the defense and adapts?

\subsection{Setup}
\label{sec:exp-setup}

We follow the protocol of Tong et al.~\cite{tong2026tokenizer}. The target
tokenizer is a BPE tokenizer trained on website-level datasets drawn from English
C4~\cite{raffel2020exploring}, using URL provenance to bucket examples by
\texttt{urlparse(url).netloc}~\cite{dodge2021documenting}. The benchmark contains
4{,}133 sites, five vocabulary sizes
\(\{80\mathrm{k},110\mathrm{k},140\mathrm{k},170\mathrm{k},200\mathrm{k}\}\),
and a 50/50 member and non-member split. Shadow tokenizers are trained with the
same BPE procedure as the target tokenizer, on random 50\% subsamples of the pool.

We report AUC, balanced accuracy, and TPR at 1\% FPR. The first two measure the
overall ranking quality of the attack. The last measures the low false-positive
regime, which is important for membership inference and often harder to improve. The distinction between ranking quality and threshold-dependent operational effectiveness is also emphasized in deployment-aware security evaluations \cite{akinrele2026prompt}.

Before evaluating our method, we reproduce the baseline's cheapest published
cell, Compression Rate at 80k vocabulary, which requires no shadow tokenizers.
Our reproduction matches the published value within the baseline's noise band:
AUC \(0.5147\) versus \(0.5070\), balanced accuracy \(0.5149\) versus \(0.5130\),
and TPR@1\%FPR \(0.82\%\) versus \(0.72\%\). This gives a basic check that our
data collection and tokenizer training pipeline matches the published setting.

\subsection{Undefended tokenizers}
\label{sec:exp-undefended}

\begin{figure}[t]
    \centering
    \includegraphics[width=\linewidth]{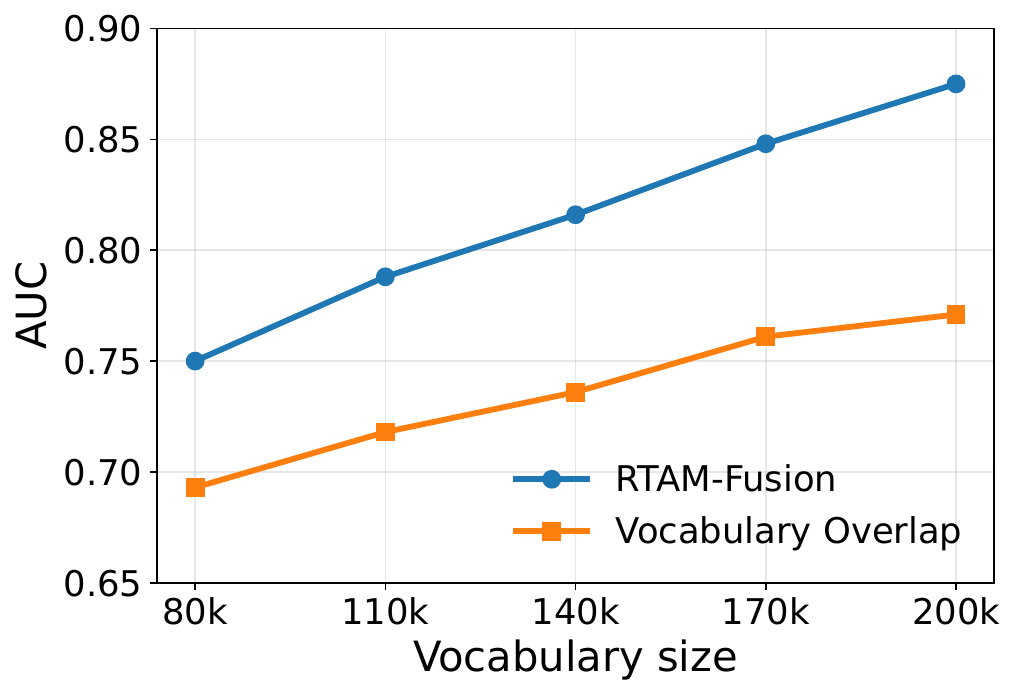}
    \caption{Undefended comparison between \textsc{RTAM-Fusion} and Vocabulary
    Overlap across vocabulary sizes. The figure shows AUC as vocabulary size
    increases.}
    \label{fig:main-undefended-curves}
\end{figure}

We first compare \textsc{RTAM-Fusion} against Vocabulary Overlap, the strongest
published baseline in Tong et al.~\cite{tong2026tokenizer}. Table~\ref{tab:undefended-surface} reports the comparison across all five
vocabulary sizes, and Figure~\ref{fig:main-undefended-curves} shows the AUC
trend visually. \textsc{RTAM-Fusion} uses the fixed fusion rule from
Section~\ref{sec:method}, with weights locked on a held-out split before
evaluation.

\begin{table}[t]
\centering
\caption{Undefended comparison between \textsc{RTAM-Fusion} and Vocabulary
Overlap across vocabulary sizes. Each metric is reported as
\textsc{RTAM-Fusion}/Vocabulary Overlap. Check marks indicate cells where
\textsc{RTAM-Fusion} is higher.}
\label{tab:undefended-surface}
\resizebox{\linewidth}{!}{%
\begin{tabular}{@{}lccc@{}}
\toprule
Vocab & AUC & BA & TPR@1\%FPR \\
\midrule
80k  & \textbf{0.750}/0.693 $\checkmark$ & \textbf{0.687}/0.666 $\checkmark$ & 23.5\%/\textbf{26.8}\% $\times$ \\
110k & \textbf{0.788}/0.718 $\checkmark$ & \textbf{0.718}/0.672 $\checkmark$ & 25.6\%/\textbf{28.8}\% $\times$ \\
140k & \textbf{0.816}/0.736 $\checkmark$ & \textbf{0.750}/0.696 $\checkmark$ & 27.0\%/\textbf{29.7}\% $\times$ \\
170k & \textbf{0.848}/0.761 $\checkmark$ & \textbf{0.782}/0.709 $\checkmark$ & \textbf{32.8}\%/32.5\% $\checkmark$ \\
200k & \textbf{0.875}/0.771 $\checkmark$ & \textbf{0.804}/0.711 $\checkmark$ & \textbf{36.8}\%/34.6\% $\checkmark$ \\
\bottomrule
\end{tabular}%
}
\end{table}

The main pattern is clear in both Table~\ref{tab:undefended-surface} and
Figure~\ref{fig:main-undefended-curves}. \textsc{RTAM-Fusion} improves AUC and
balanced accuracy at every vocabulary size. The AUC gains range from about
\(+0.06\) to \(+0.10\), showing that rare-token activation mass gives a stronger
overall ranking signal than raw distinctive-token overlap. The 200-split
robustness check gives the same conclusion: \textsc{RTAM-Fusion} wins AUC in
every vocabulary cell and wins balanced accuracy in 98 to 100\% of split samples.

The low false-positive regime is more mixed. \textsc{RTAM-Fusion} wins
TPR@1\%FPR at the larger vocabulary sizes, especially 170k and 200k, but loses at
the smallest vocabulary sizes. This is an important boundary of the method rather
than a reporting nuisance. Near-unique rare tokens are most useful for the
low-FPR tail, and small vocabularies leave fewer such tokens available. The
200-split check confirms this vocabulary dependence, with TPR@1\%FPR win rate
ranging from 2\% at 80k to 100\% at 200k.

\subsection{Min-count defense}
\label{sec:exp-defense}

We next evaluate the baseline's min-count defense. This defense removes tokens
whose aggregate training count is below a threshold \(n_{\min}\). It directly
targets the rare tokens that \textsc{RTAM-Fusion} uses, so it is a useful stress
test of the method.

For this experiment, we use defense-aware re-locking. The feature family is
unchanged, but the fusion weights are selected on a held-out calibration split
from the defended setting. This models an attacker who knows that the defense was
applied.

\begin{table*}[t]
\centering
\caption{Min-count defense comparison between \textsc{RTAM-Fusion} and
Vocabulary Overlap across vocabulary sizes and pruning thresholds. Each cell
reports AUC as \textsc{RTAM-Fusion}/Vocabulary Overlap.}
\label{tab:defense-surface}

\begin{tabular}{@{}lccccc@{}}
\toprule
$n_{\min}$ & 80k & 110k & 140k & 170k & 200k \\
\midrule
32 & \textbf{0.823}/0.663 & \textbf{0.860}/0.699 & \textbf{0.880}/0.718 & \textbf{0.878}/0.736 & \textbf{0.877}/0.746 \\
48 & \textbf{0.825}/0.663 & \textbf{0.843}/0.697 & \textbf{0.825}/0.714 & \textbf{0.822}/0.717 & \textbf{0.820}/0.734 \\
64 & \textbf{0.822}/0.663 & \textbf{0.788}/0.685 & \textbf{0.770}/0.699 & \textbf{0.768}/0.707 & \textbf{0.768}/0.717 \\
\bottomrule
\end{tabular}
\end{table*}

Table~\ref{tab:defense-surface} shows that \textsc{RTAM-Fusion} still improves
over Vocabulary Overlap on AUC and balanced accuracy across all published
min-count thresholds and vocabulary sizes. The strongest defended 200k setting
with \(n_{\min}=64\) gives AUC \(0.768\) for \textsc{RTAM-Fusion} versus
\(0.717\) for Vocabulary Overlap. The TPR@1\%FPR result changes. Under min-count pruning, \textsc{RTAM-Fusion}
loses the low-FPR comparison, with defended TPR around 13 to 16\% versus 22 to
31\% for Vocabulary Overlap. This is expected. The defense deletes many of the
near-unique tokens that drive the low-FPR tail. The broader rarity features,
especially \(\mathrm{meanidf}\) and larger rare-mass cutoffs, still preserve the
ranking signal, but the most extreme rare-token evidence is weakened. A size-conditioned version of this experiment is given in
Appendix~\ref{app:size-defense}. The same pattern holds across dataset-size
bins: \textsc{RTAM-Fusion} improves AUC and balanced accuracy, while the
low-FPR tail remains harder.

\subsection{Shadow-tokenizer efficiency}
\label{sec:exp-efficiency}

The attack also needs fewer shadow tokenizers than Vocabulary Overlap.
\textsc{RTAM-Fusion} uses between 8 and 20 shadow tokenizers depending on the
setting, while Vocabulary Overlap uses 96 in the baseline protocol
\cite{tong2026tokenizer}. Since shadow-tokenizer training dominates the cost,
we treat shadow count as part of the result. We do not report a new wall-clock
comparison because our runs were performed across remote CPU and Kaggle
environments rather than a controlled timing setup. Instead, we use the
baseline's own cost table as a reference and report the conservative claim:
\textsc{RTAM-Fusion} reaches comparable or better AUC and balanced accuracy with
roughly 5 to 12 times fewer shadow tokenizers.

\subsection{Adaptive defenses}
\label{sec:exp-adaptive}

Finally, we evaluate defenses against adaptive attackers. A static evaluation
asks whether a fixed attack fails after a defense is applied. An adaptive
evaluation asks whether the membership signal is still present once the attacker
knows the defense and changes its statistic. This distinction matters because a
defense may remove one feature without removing the underlying signal.

\begin{figure}[t]
    \centering
    \includegraphics[width=\linewidth]{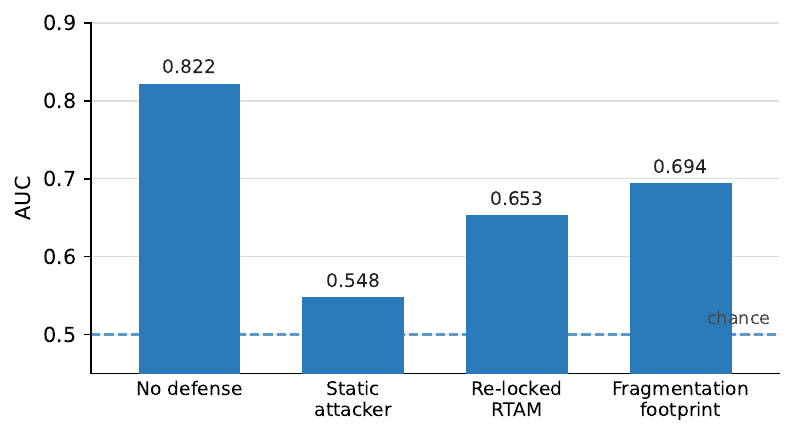}
    \caption{Adaptive attacks against targeted token removal. The defense weakens
    a naive \textsc{RTAM-Fusion} attacker, but defense-aware attacks recover
    membership signal.}
    \label{fig:adaptive-targeted-arc}
\end{figure}

Figure~\ref{fig:adaptive-targeted-arc} shows this effect for targeted token
removal. The pattern illustrates why we treat adaptive evaluation as the relevant
defense test, rather than as an optional stress test. The defense weakens a naive
attacker, but the signal returns once the attacker adapts. Table~\ref{tab:adaptive-summary}
summarizes the broader production-vocabulary results. Once the attacker adapts,
every production-vocabulary defense remains above chance. Targeted token removal
and min-count pruning both leave recoverable signal, while merge-order
randomization is nearly flat across the temperature sweep. Full per-attack
details and privacy-utility frontiers are given in
Appendices~\ref{app:adaptive-arc} and~\ref{app:frontier}.

\begin{table}[t]
\centering
\caption{Adaptive-defense results. The table reports the strongest adaptive
attack for each defense, with production-vocabulary results except where a
smaller vocabulary is explicitly named.}
\label{tab:adaptive-summary}
\resizebox{\linewidth}{!}{%
\begin{tabular}{@{}lcc@{}}
\toprule
Defense & Static or initial result & Best adaptive AUC \\
\midrule
Targeted token removal & 0.548 & 0.694 \\
Min-count, $n_{\min}=64$ & 0.609 & 0.761 \\
Min-count, $n_{\min}=256$ & stronger pruning & $\approx 0.591$ \\
Merge-order randomization & order randomized & 0.710--0.712 \\
Bounded contribution & 0.523 at 12k vocab & 0.725 at 200k vocab \\
\bottomrule
\end{tabular}%
}
\end{table}

Targeted token removal looks effective against a naive attacker, reducing AUC to
\(0.548\), but adaptive attacks recover AUC \(0.653\) with re-locked RTAM
features and \(0.694\) with a fragmentation-footprint statistic. Min-count
pruning also leaks adaptively: at \(n_{\min}=64\), the best adaptive attack
reaches AUC \(0.761\), and stronger pruning lowers this to about \(0.591\) only
after removing roughly 115{,}000 tokens. Merge-order randomization is nearly
flat, with adaptive AUC between \(0.710\) and \(0.712\), matching
Proposition~\ref{prop:ancillary}. Bounded contribution succeeds at 12k
vocabulary with AUC \(0.523\), but rises to \(0.725\) at production vocabulary
size. Overall, none of the tested defenses drives adaptive membership inference
to chance at production scale. They either remove the wrong channel, leave a
footprint, or look stronger at small vocabulary size than at production size.

\section{Conclusion}
\label{sec:conclusion}

Tokenizers are not neutral preprocessing tools. They are learned from text, and
their vocabularies can retain traces of the corpora that trained them. This paper
showed that, for BPE tokenizer membership inference, the strongest signal comes
from rare tokens activated by a candidate dataset. Merge order and merge-graph
structure are not where the main membership evidence lives. This observation led
to \textsc{RTAM-Fusion}, a calibrated rare-token activation mass attack that
improves AUC and balanced accuracy over the strongest published baseline while
using fewer shadow tokenizers. The low false-positive regime is more mixed, with
the clearest gains at production vocabulary sizes. The defense results tell the
same story from the other side. Randomizing merge order does little because order
is not the channel. Removing rare tokens can weaken the original signal, but it
can also leave a fragmentation footprint that an adaptive attacker can read. Our
study is limited to BPE tokenizers and the site-level C4 benchmark, so
cross-tokenizer, out-of-distribution, and canary-style evaluations remain
important next steps. Tokenizer tokens tell
training tales, even when weights are never queried.

\section*{Limitations}

This study focuses on BPE tokenizers and on the site-level C4 benchmark used in prior tokenizer membership work. The threshold model in Section~\ref{sec:theory} is an explanatory model for this setting, not a complete account of tokenizer training. Real BPE changes the alphabet as merges are learned, and token inclusion events are dependent because tokens compete for a fixed vocabulary budget. Our results should therefore be read as evidence that rare-token existence is the main membership channel in this benchmark, not as a finite-sample optimality claim for every tokenizer. Several evaluations remain open, including other tokenizer families such as WordPiece, unigram tokenizers, and SentencePiece variants, along with out-of-distribution, cross-corpus, and canary-style tests. Our defense results are limited to min-count pruning, merge-order randomization, targeted token removal, and bounded per-site contribution. None drives adaptive membership inference to chance at production vocabulary size, but this does not show that tokenizer defenses are impossible. We also do not evaluate content-level differentially private tokenizer training, where noise is applied to token counts during learning rather than after the vocabulary has already been built. Finally, our attack assumes that the adversary can train shadow tokenizers from a similar broad distribution and knows the tokenizer training procedure, matching the prior benchmark but not every deployment setting.

\bibliography{custom}

\appendix

\section{Additional Method Details}
\label{app:method-extra}

\noindent \textbf{Lineage.}
The threat model and shadow-tokenizer protocol are inherited from Tong
et al.~\cite{tong2026tokenizer}. The calibration idea follows LiRA
\cite{carlini2022membership}. Our contribution is the rare-token activation mass
statistic, its multi-level fusion, and the defense-aware re-locking used in the
adaptive evaluation.

\noindent \textbf{Detailed cost.}
Computing all rare-mass and IDF features is linear in the amount of tokenized
text once the shadow tokenizers are available. The calibration step is $O(N)$ per
candidate dataset given the shadow features. The main cost is therefore the
number of shadow tokenizers. Using the baseline's own reported training costs,
RTAM-Fusion's shadow budget is substantially smaller than the $N=96$ setting used
for Vocabulary Overlap~\cite{tong2026tokenizer}.

\noindent \textbf{Algorithm.} Algorithm~\ref{alg:rtam-fusion} gives the scoring procedure used by
\textsc{RTAM-Fusion}. The shadow tokenizers are assumed to have already been
trained under the threat model in Section~\ref{sec:method-threat}. Given those
tokenizers, the algorithm computes the same features defined in
Section~\ref{sec:method}: rare-token activation mass at each cutoff, the two
IDF-style rarity aggregates, and their calibrated versions under the out-shadow
null. The final output is the fused membership score thresholded at \(\tau\).

\begin{algorithm}[t]
\caption{\textsc{RTAM-Fusion} membership inference}
\label{alg:rtam-fusion}
\begin{algorithmic}[1]
\Require Candidate dataset $D$, target vocabulary $V_{\mathrm{target}}$,
shadow tokenizers $\{V_s\}_{s=1}^N$, rarity cutoffs $\mathcal{K}$, threshold $\tau$
\For{each tokenizer $X \in \{V_{\mathrm{target}}\} \cup \{V_s\}_{s=1}^N$}
    \State Tokenize $D$ with $X$ and compute token counts $\mathrm{cnt}_D(t;X)$
    \State Estimate shadow frequency $\mathrm{sf}(t)$ for each activated token
    \State Compute $\mathrm{rm}_K(D;X)$ for each $K \in \mathcal{K}$
    \State Compute $\mathrm{meanidf}(D;X)$ and $\mathrm{maxidf}(D;X)$
\EndFor
\For{each feature $\phi$}
    \State Build the out-shadow null $\mathrm{OUT}_{\phi}(D)$
    \State Calibrate $\phi(D;V_{\mathrm{target}})$ against this null
\EndFor
\State Fuse the calibrated features into $S(D)$
\State \Return $\mathbb{1}[S(D)>\tau]$
\end{algorithmic}
\end{algorithm}

\section{Additional Experimental Results}
\label{app:additional-experiments}

\subsection{Published baseline surface}
\label{app:baseline-surface}

Table~\ref{tab:baseline-surface} reproduces the published attack surface of
Tong et al.~\cite{tong2026tokenizer}. We use these values as the literature
comparators for Vocabulary Overlap, Frequency Estimation, Naive Bayes, Merge
Similarity, and Compression Rate.

\begin{table*}[t]
\centering
\small
\caption{Published tokenizer-membership baselines from Tong et al.~\cite{tong2026tokenizer},
Table 3. Vocabulary Overlap (VO) is the strongest baseline across vocabulary
sizes and metrics, with Frequency Estimation (FE) the next strongest.}
\label{tab:baseline-surface}
\begin{tabular}{l l r r r r r}
\toprule
Attack & Metric & 80k & 110k & 140k & 170k & 200k \\
\midrule
\multirow{3}{*}{CR}
 & AUC        & 0.507 & 0.509 & 0.508 & 0.508 & 0.509 \\
 & BA         & 0.513 & 0.517 & 0.514 & 0.513 & 0.516 \\
 & TPR@1\%FPR & 0.72\% & 0.71\% & 0.68\% & 0.82\% & 0.77\% \\
\midrule
\multirow{3}{*}{NB ($k{=}20$k)}
 & AUC        & 0.534 & 0.526 & 0.535 & 0.546 & 0.564 \\
 & BA         & 0.526 & 0.524 & 0.533 & 0.538 & 0.551 \\
 & TPR@1\%FPR & 3.78\% & 3.44\% & 4.11\% & 5.86\% & 8.03\% \\
\midrule
\multirow{3}{*}{MS}
 & AUC        & 0.493 & 0.494 & 0.494 & 0.495 & 0.495 \\
 & BA         & 0.509 & 0.507 & 0.508 & 0.506 & 0.508 \\
 & TPR@1\%FPR & 1.06\% & 1.02\% & 0.97\% & 0.92\% & 0.87\% \\
\midrule
\multirow{3}{*}{FE}
 & AUC        & 0.610 & 0.645 & 0.676 & 0.707 & 0.740 \\
 & BA         & 0.614 & 0.641 & 0.660 & 0.681 & 0.705 \\
 & TPR@1\%FPR & 21.30\% & 22.00\% & 22.41\% & 25.61\% & 27.88\% \\
\midrule
\multirow{3}{*}{\textbf{VO}}
 & AUC        & \textbf{0.693} & \textbf{0.718} & \textbf{0.736} & \textbf{0.761} & \textbf{0.771} \\
 & BA         & \textbf{0.666} & \textbf{0.672} & \textbf{0.696} & \textbf{0.709} & \textbf{0.711} \\
 & TPR@1\%FPR & \textbf{26.77\%} & \textbf{28.75\%} & \textbf{29.72\%} & \textbf{32.53\%} & \textbf{34.61\%} \\
\bottomrule
\end{tabular}
\end{table*}

\begin{figure}[t]
    \centering
    \includegraphics[width=\linewidth]{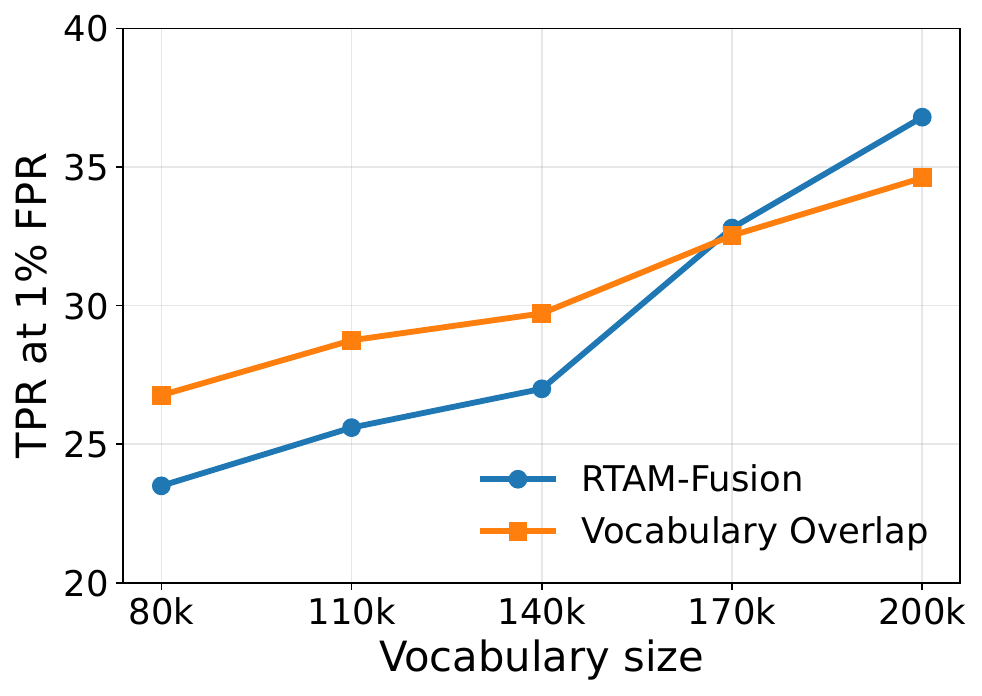}
    \caption{Low false-positive comparison between \textsc{RTAM-Fusion} and
    Vocabulary Overlap across vocabulary sizes. The figure reports TPR at
    1\% FPR in the undefended setting.}
    \label{fig:appendix-tpr-vocab}
\end{figure}

Figure~\ref{fig:appendix-tpr-vocab} shows the low-FPR comparison separately,
highlighting that the TPR@1\%FPR result is vocabulary-dependent.

\subsection{Size-conditioned defense surface}
\label{app:size-defense}

We also compare against the baseline's size-conditioned min-count defense
surface. Datasets are bucketed by size into \([0,400)\), \([400,800)\), and
\([800,1200]\), following Tong et al.~\cite{tong2026tokenizer}. Table~\ref{tab:size-defense-surface}
shows that \textsc{RTAM-Fusion} improves AUC and balanced accuracy across all
size and vocabulary cells. The smallest margin is the baseline's headline
large-dataset 200k cell, where \textsc{RTAM-Fusion} reaches \(0.805\) against
\(0.797\).

\begin{table}[t]
\centering
\small
\caption{Size-conditioned min-count defense results at $n_{\min}{=}64$.
Each cell reports RTAM-Fusion/Vocabulary Overlap AUC, with datasets grouped by
candidate dataset size. Vocabulary Overlap values are from Tong
et al.~\cite{tong2026tokenizer}, Table 9.}
\label{tab:size-defense-surface}
\begin{tabular}{l r r r}
\toprule
Vocab & $[0,400)$ & $[400,800)$ & $[800,1200]$ \\
\midrule
80k  & \textbf{0.730}/0.646 & \textbf{0.786}/0.703 & \textbf{0.782}/0.712 \\
110k & \textbf{0.779}/0.662 & \textbf{0.846}/0.731 & \textbf{0.827}/0.743 \\
140k & \textbf{0.816}/0.677 & \textbf{0.881}/0.751 & \textbf{0.868}/0.768 \\
170k & \textbf{0.799}/0.683 & \textbf{0.847}/0.761 & \textbf{0.846}/0.795 \\
200k & \textbf{0.751}/0.694 & \textbf{0.819}/0.772 & \textbf{0.805}/0.797 \\
\bottomrule
\end{tabular}

\vspace{4pt}
\footnotesize Bold indicates the higher AUC within each RTAM-Fusion/Vocabulary
Overlap comparison.
\end{table}

\begin{figure}[t]
    \centering
    \includegraphics[width=\linewidth]{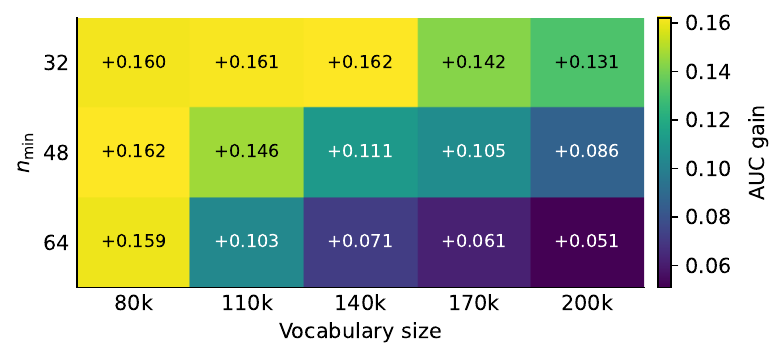}
    \caption{AUC gain of \textsc{RTAM-Fusion} over Vocabulary Overlap under the
    min-count defense. Each cell reports the difference in AUC for a pruning
    threshold and vocabulary size.}
    \label{fig:appendix-mincount-gain}
\end{figure}

\begin{figure}[t]
    \centering
    \includegraphics[width=\linewidth]{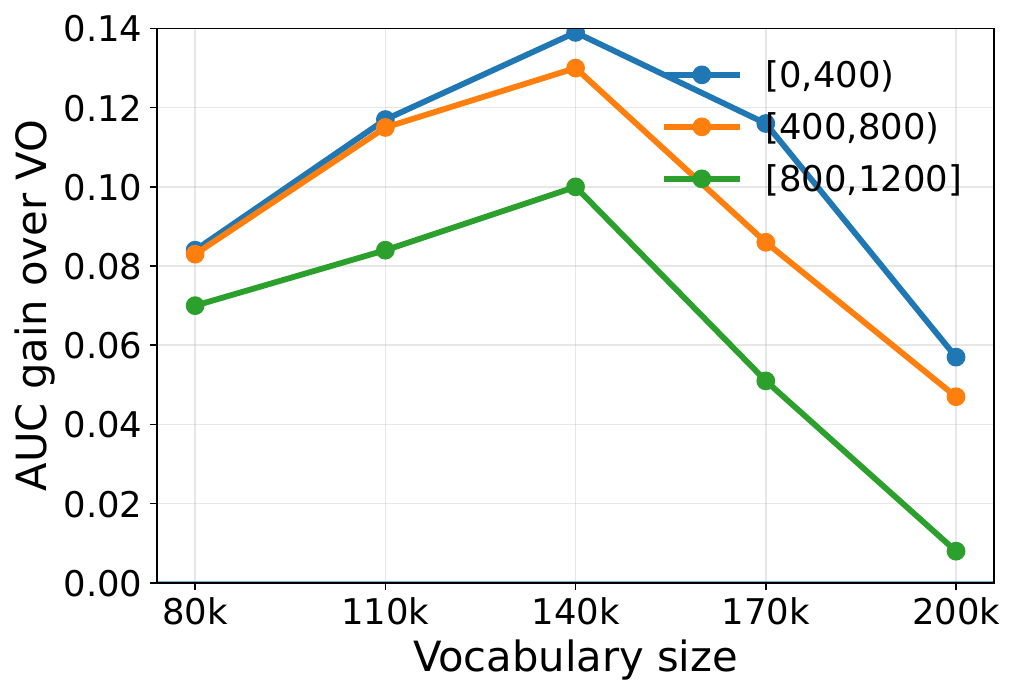}
    \caption{Size-conditioned AUC gain of \textsc{RTAM-Fusion} over Vocabulary
    Overlap under the min-count defense at \(n_{\min}=64\). Candidate datasets
    are grouped by size.}
    \label{fig:appendix-size-gain}
\end{figure}

Figures~\ref{fig:appendix-mincount-gain} and~\ref{fig:appendix-size-gain}
summarize the defended AUC gains across pruning thresholds, vocabulary sizes,
and dataset-size bins.

\subsection{Adaptive-defense details}
\label{app:adaptive-arc}

Table~\ref{tab:adaptive-arc} gives the per-attack details behind the adaptive
evaluation in Section~\ref{sec:exp-adaptive}. It reports the static or naive
result where available, together with the strongest adaptive attack after the
defense is known.

\begin{table*}[t]
\centering
\small
\caption{Adaptive attacks against targeted token removal. A naive RTAM-Fusion
attacker is weakened by the defense, while defense-aware attacks recover
membership signal. The undefended RTAM-Fusion AUC is $0.822$.}
\label{tab:adaptive-arc}
\begin{tabular}{l l r}
\toprule
Attack & Description & AUC \\
\midrule
A0 & Static RTAM-Fusion & 0.548 \\
A1 & Re-locked rarity feature & 0.653 \\
A2 & Compression-based fragmentation footprint & \textbf{0.694} \\
\bottomrule
\end{tabular}
\end{table*}

\subsection{Adaptive privacy and utility frontier}
\label{app:frontier}

Table~\ref{tab:defense-frontier} reports the production-vocabulary
privacy-utility frontier for all defense settings we evaluate. The table includes
the utility cost of each setting, measured in bytes per token where available.
No tested setting reaches chance-level adaptive AUC at production vocabulary
size.

\begin{table*}[t]
\centering
\small
\caption{Adaptive-defense frontier at 200k vocabulary size. Each row reports
the strongest adaptive attack observed for a defended tokenizer, together with
the available utility measurement.}
\label{tab:defense-frontier}
\begin{tabular}{l r r l}
\toprule
Defense mechanism & Adaptive AUC & Utility cost & Symmetric? \\
\midrule
Targeted token removal, $K{=}32$k & 0.694 & $-0.27\%$ bpt & No \\
Min-count, $n_{\min}{=}256$ & 0.591 & 115k tokens removed & Yes \\
Merge-order randomization & 0.710--0.712 & not measured at matched AUC & Yes \\
Bounded-contribution BPE, $C{=}1$ & 0.725 & $-1.2\%$ bpt at 12k vocab & Yes \\
\bottomrule
\end{tabular}
\end{table*}

\begin{figure}[t]
    \centering
    \includegraphics[width=\linewidth]{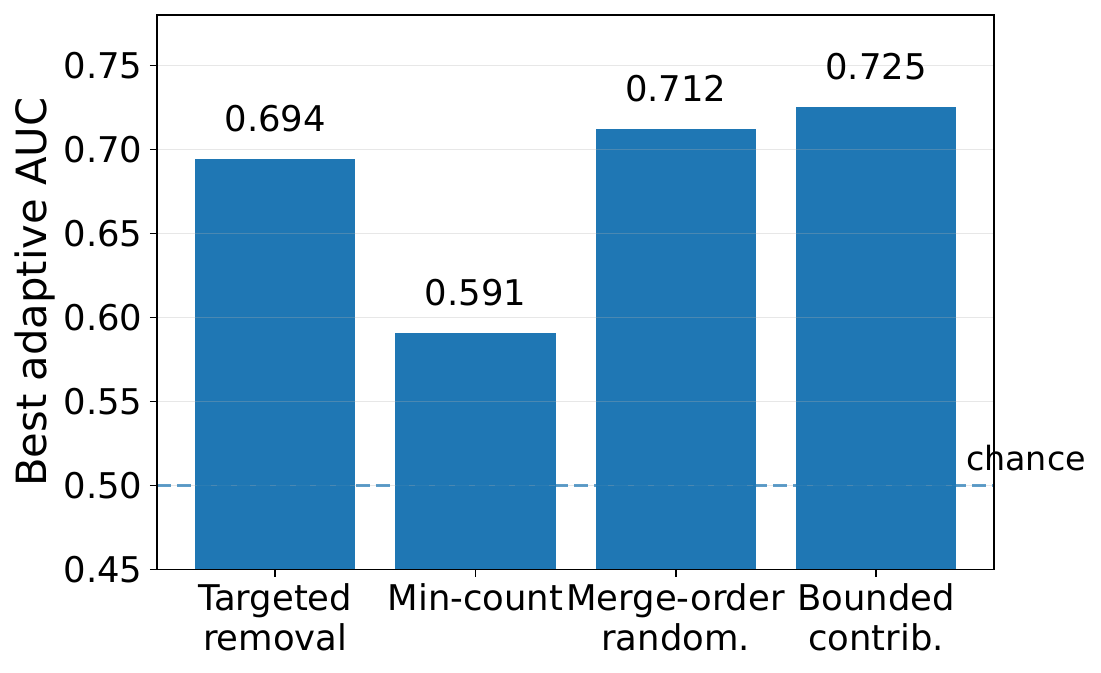}
    \caption{Adaptive-defense comparison at 200k vocabulary size. Each bar shows
    the strongest adaptive attack observed for a defended tokenizer.}
    \label{fig:appendix-adaptive-defense-comparison}
\end{figure}

Figure~\ref{fig:appendix-adaptive-defense-comparison} visualizes the same
frontier, showing that all tested production-vocabulary defenses remain above
chance under adaptive attack.

\subsection{Exploratory search for non-existence signals}
\label{app:noble-search}

We also ran an exploratory search for statistics outside the existence and
rarity family used by \textsc{RTAM-Fusion}. The search considered candidate
signals inspired by several areas, including topology, random matrix theory,
percolation, dynamical systems, scale-space analysis, and extreme-value theory.
Most candidates reduced, after inspection, to either existence-like signals or
order-like signals. The one genuinely different candidate was a second-order
co-occurrence statistic. It reached standalone AUC \(0.574\) and added only
\(+0.008\) AUC when fused with \textsc{RTAM-Fusion}. We treat this as an
exploratory negative result rather than as evidence of a formal sufficiency
theorem.

\section{Theory Details and Scope}
\label{app:theory-proofs}

This appendix gives the details behind Section~\ref{sec:theory}. The goal is not
to provide an exact finite-sample model of BPE training. Instead, we make explicit
the assumptions under which the threshold argument holds, and separate the formal
likelihood-ratio result from the more approximate claim about merge order.

\subsection{Threshold model and assumptions}
\label{app:threshold-model}

We use the threshold model from Section~\ref{sec:theory-setup}. A token $t$ is
included in the final vocabulary if its corpus frequency exceeds the threshold
induced by the vocabulary size:
\[
t \in V
\quad \Longleftrightarrow \quad
\mathrm{freq}_C(t) \geq \theta_{|V|}.
\]
For a candidate dataset $D$, corpus frequency decomposes as
\[
\mathrm{freq}_{C}(t)
=
\mathrm{freq}_{C \setminus D}(t)
+
\mathbb{1}[D \subseteq C]\mathrm{freq}_{D}(t).
\]
We treat $\mathrm{freq}_{C \setminus D}(t)$ as an unknown background quantity
whose distribution is shared under membership and non-membership. This is the
nuisance term estimated indirectly by shadow tokenizers.

The model makes two simplifying assumptions. First, it treats token inclusion as
a threshold on final token frequency, while real BPE changes the alphabet during
training. Second, it treats token-existence observations as conditionally
independent given membership. In real tokenizers, tokens compete for a fixed
vocabulary budget, so the observations are not exactly independent. These
assumptions are used to obtain a tractable likelihood-ratio characterization.

\subsection{Derivation of Theorem~\ref{thm:np-optimal}}
\label{app:thm-np}

Let $e_i=\mathbb{1}[t_i \in V]$ be the existence indicator for a candidate token
activated by $D$. Under the conditional-independence assumption,
\[
P(\{e_i\}_i \mid H)
=
\prod_i P(e_i \mid H),
\qquad
H \in \{H_0,H_1\}.
\]
The Neyman--Pearson lemma states that, for testing $H_1$ against $H_0$ at a fixed
false-positive rate, the most powerful test thresholds the likelihood ratio.
Taking logs gives
\[
\Lambda(D)
=
\sum_i
\log
\frac{P(e_i \mid H_1)}{P(e_i \mid H_0)}.
\]

It remains to see why the largest terms correspond to rare tokens that are
frequent in $D$. Let
\[
B_i = \mathrm{freq}_{C \setminus D}(t_i)
\]
be the background frequency of token $t_i$. Under non-membership,
\[
P(e_i=1 \mid H_0)
=
P(B_i \geq \theta_{|V|}).
\]
Under membership,
\[
P(e_i=1 \mid H_1)
=
P(B_i + \mathrm{freq}_{D}(t_i) \geq \theta_{|V|})
\]
\[
P(e_i=1 \mid H_1)
=
P(B_i \geq \theta_{|V|}-\mathrm{freq}_{D}(t_i))
\]
Thus the gap between the two probabilities increases with
$\mathrm{freq}_{D}(t_i)$. If $D$ uses a token many times, membership can move the
token across the threshold.

The gap also depends on the background frequency. If $B_i$ is already large, then
both probabilities are close to one, and observing the token gives little
membership evidence. If $B_i$ is far below the threshold, then both probabilities
are small unless $D$ supplies enough mass to move the token across the threshold.
The largest likelihood-ratio terms therefore occur for tokens that are rare in
the background but frequent in $D$.

The exact likelihood ratio depends on the unknown distribution of $B_i$ for each
token. \textsc{RTAM-Fusion} approximates this structure with shadow tokenizers:
shadow frequency estimates token rarity, and the out-shadow calibration estimates
how unusual the target statistic is under non-membership. This is why the method
uses rarity-weighted activated-token mass rather than raw overlap alone.

\subsection{Order and graph features}
\label{app:order-graph}

Proposition~\ref{prop:ancillary} is a model implication, not an exact theorem for
all finite BPE runs. The intuition is as follows. Merge ranks are determined
mostly by the global frequency ordering of the training corpus. For a single
candidate dataset $D$, the main membership-relevant effect is that $D$ may push
some rare tokens across the inclusion threshold. Once we condition on which tokens
exist and how often $D$ uses them, the remaining merge-order information mostly
reflects the background corpus.

A simple way to express this is to decompose a token's merge rank into a
background component and a perturbation from $D$:
\[
R(t)
=
\rho(C \setminus D, t)
+
\delta_D(t).
\]
Here $\rho(C \setminus D, t)$ is the rank contribution induced by the background
corpus, while $\delta_D(t)$ is the perturbation caused by including $D$. Under the
large-corpus threshold view, $\delta_D(t)$ matters mainly when $D$ changes whether
the token crosses the inclusion threshold. That event is already captured by the
existence vector $E$ and by $D$'s own frequency $f_D$. The remaining background
ordering has the same distribution under $H_0$ and $H_1$.

This gives the conditional-independence statement used in the main text:
\[
P(\mathrm{member} \mid E, f_D, R, G)
=
P(\mathrm{member} \mid E, f_D).
\]
The same reasoning applies to merge-graph position $G$, since the graph is built
from the same merge events. The approximation may fail if one candidate dataset is
large enough to substantially reorder many high-frequency merges. This is not the
regime studied in our experiments, where candidates are individual site-level
datasets drawn from a larger pool.

\subsection{Implications for defenses}
\label{app:defense-implications}

The theory suggests two expectations for defenses. First, a defense that changes
only merge order should not remove the main membership channel. The likelihood
ratio in Theorem~\ref{thm:np-optimal} depends on token existence and rarity, not
on the order in which surviving tokens were created.

Second, defenses that remove vocabulary content can reduce the original
rare-token signal, but they may create a different signal. If a member dataset
previously used a distinctive token and that token is removed, the dataset must be
re-tokenized into smaller pieces. This can change its tokenization pattern more
than it changes the pattern of a non-member dataset that did not rely on the
removed token. We call this a fragmentation footprint.

We do not formalize this footprint as a theorem. Doing so would require a model
of how BPE re-tokenizes strings after token removal, together with a separate
detection-theoretic analysis for the footprint statistic. Instead, we treat it as
a mechanism suggested by the theory and tested empirically in the adaptive-defense
experiments.

\subsection{Scope}
\label{app:theory-scope}

The theory supports three claims. First, under the threshold model, the optimal
existence-based membership test gives the largest weight to tokens that are rare
in the background and frequent in the candidate dataset. Second, RTAM-style
features are calibrated, finite-sample approximations to this likelihood-ratio
structure. Third, merge order and merge-graph features are not expected to add
membership signal once token existence and candidate-token frequency are known.

The theory does not claim that \textsc{RTAM-Fusion} is exactly optimal for finite
BPE training. It also does not claim that no tokenizer defense can work. Real BPE
uses an evolving vocabulary, token-existence events are dependent, and
content-removal defenses require a separate model of re-tokenization after
pruning. We use the theory as a characterization of the main membership channel
and as a guide for the adaptive evaluations, not as a complete impossibility
result.

\end{document}